\documentclass[twocolumn,english,aps,prb,twocolum,superscriptaddress,natbib,bibnotes,amsmath,amssymb,floatfix,groupedaddress,footinbib]{revtex4-2}

\usepackage[markup=nocolor, authormarkupposition=left]{changes} 

\usepackage{soul}
\usepackage[utf8]{inputenc}
\usepackage[english]{babel}
\usepackage{amsmath,amsfonts,amssymb}
\usepackage[T1]{fontenc}
\usepackage{url}

\usepackage{afterpage}

\usepackage{amsmath}
\usepackage{siunitx}
\usepackage[version=4]{mhchem}
\usepackage{amsfonts}
\usepackage{amssymb}

\usepackage[colorlinks=true,citecolor=blue,linkcolor=magenta]{hyperref}

\usepackage{epstopdf}
\usepackage{graphicx}
\graphicspath{{./Figures/}}

\usepackage{changes}

\begin{document}

\title{Training deep physical neural networks with local physical information bottleneck}

\author{
  Hao~Wang$^{1,7\dagger}$,
	Ziao~Wang$^{2\dagger}$,
    Xiangpeng~Liang$^{3}$,
    Han~Zhao$^{3}$,
    Jianqi~Hu$^{4}$,
    Junjie~Jiang$^{1}$,
    Xing~Fu$^{1,5,6}$,
    Jianshi~Tang$^{3}$,
    Huaqiang~Wu$^{3}$,
	  Sylvain~Gigan$^{2\ast}$,
    and Qiang~Liu$^{1,5,6\ast}$
}
\affiliation{
$^1$Department of Precision Instrument, Tsinghua University, Beijing 100084, China.\\
$^2$Laboratoire Kastler Brossel, École Normale Supérieure - Paris Sciences et Lettres (PSL) Research University, Sorbonne Université, Centre National de la Recherche Scientifique (CNRS), UMR 8552, Collège de France, Paris 75005, France.\\
$^3$School of Integrated Circuits, Beijing Advanced Innovation Center for Integrated Circuits, BNRist, Tsinghua University, Beijing 100084, China.\\
$^4$Department of Electrical and Electronic Engineering, The University of Hong Kong, Hong Kong, China.\\
$^5$State Key Laboratory of Precision Space-time Information Sensing Technology, Department of Precision Instrument, Tsinghua University, Beijing 100084, China.\\
$^6$Key Laboratory of Photonic Control Technology, Ministry of Education, Tsinghua University, Beijing 100084, China.\\
$^7$Current Address: Department of Electrical Engineering, City University of Hong Kong, Hong Kong, China.
}

\maketitle

\noindent\textbf{\noindent
Deep learning has revolutionized modern society but faces growing energy and latency constraints. Deep physical neural networks (PNNs) are interconnected, in-memory computing systems that directly exploit analog dynamics for energy-efficient, ultrafast AI execution. Realizing this potential, however, requires universal training methods tailored to their physical intricacies. Here, we present the Physical Information Bottleneck (PIB), a general and efficient framework that integrates information theory and local learning, enabling deep PNNs to learn under in principle arbitrary physical dynamics. By allocating matrix-based information bottlenecks to each unit, we demonstrate supervised, unsupervised, and reinforcement learning across electronic memristive chips and optical computing platforms. PIB also adapts to severe hardware faults and allows for parallel training via geographically distributed resources. Bypassing auxiliary digital models and contrastive measurements, PIB recasts PNN training as an intrinsic, scalable information-theoretic process compatible with diverse physical substrates. 
}

\section*{Introduction} 

\noindent{Deep} neural networks (DNNs) underpin modern artificial intelligence, enabling advances that span everyday technologies to scientific discovery~\cite{carleo2019machine}. Their remarkable performance is tied to scale: increasing network depth and size leads to emergent capabilities absent in smaller models~\cite{wei2022emergent}. For decades, this scaling has been sustained by progress in digital computing hardware. Today, however, further growth of DNNs is increasingly constrained by physical limits and escalating energy costs, raising concerns about the long-term sustainability of purely digital approaches.

\begin{figure*}[!htbp]
  \centering{
  \includegraphics[width = 1.0\linewidth]{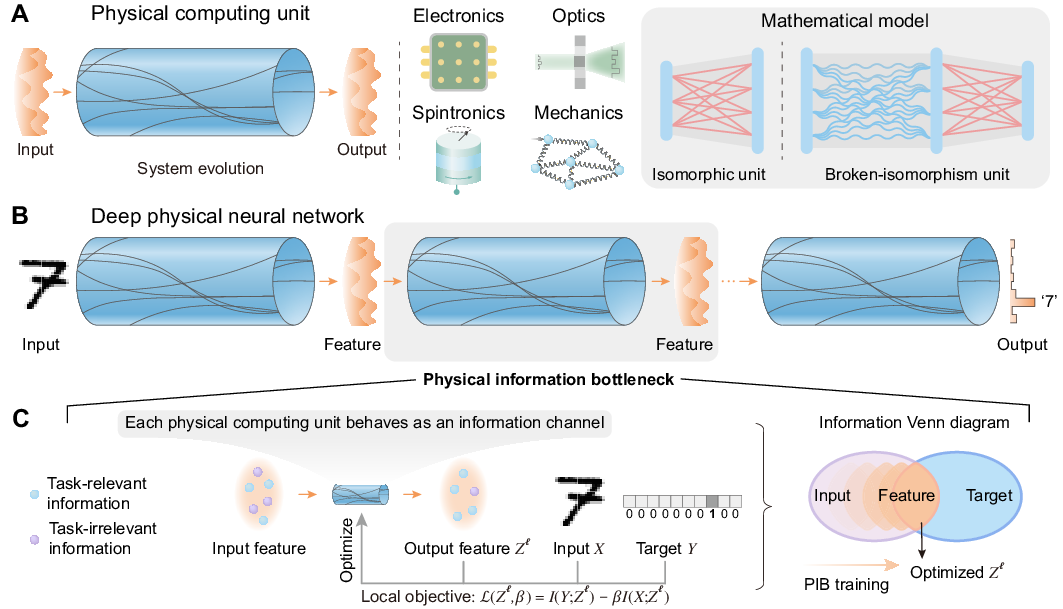}
  } 
    \caption{\noindent\textbf{Overview of deep physical neural networks (PNNs) trained with physical information bottleneck (PIB).} (\textbf{A}) A physical computing unit leverages intrinsic analog physical dynamics (indicated by illustrative trajectories) to transform an input state to an output state. These units can be built on various physical substrates, such as electronics and optics. They can be categorized into two classes based on their mathematical abstraction: isomorphic units that are designed to emulate explicit and reconfigurable mathematical operators, and broken-isomorphism (or non-isomorphic) units that exploit native, fixed physical dynamics augmented by a trainable layer. (\textbf{B}) Similar to the hierarchical architecture of digital deep neural networks (DNNs), deep PNNs are constructed by cascading individual physical computing units. (\textbf{C}) To achieve the desired functionalities, each unit is optimized locally via the PIB objective. This framework promotes the output feature of each unit to retain task-relevant information (e.g., for target prediction) while filtering out task-irrelevant information from the original input. The PIB training dynamics are schematically illustrated by the information Venn diagram (right), where the varying orange regions denote the matrix-based mutual information shared between the learned feature, the input, and the target.}.
    \label{fig1}
\end{figure*}

The economic and ecological necessities created by this asynchrony, along with fundamental scientific curiosity, have renewed interest in physical neural networks (PNNs)~\cite{abreu2024photonics}.
Partly inspired by biological brains, PNNs exploit the intrinsic analog dynamics of physical systems for computation~\cite{hopfield1982neural}, such as those built on analog electronics~\cite{sebastian2020memory}, optics~\cite{wetzstein2020inference,shastri2021photonics,zhou2022photonic}, spintronics~\cite{torrejon2017neuromorphic,grollier2020neuromorphic}, and mechanics~\cite{stern2020supervised} (Fig.~\ref{fig1}A). Recent theoretical and experimental advances~\cite{jaeger2023toward,mcmahon2023physics} have expanded these systems from standalone computing units into deep PNNs capable of hierarchical processing analogous to DNNs (Fig.~\ref{fig1}B)~\cite{wright2022deep,nakajima2022physical,momeni2023backpropagation}.
By leveraging substrate-specific physical processes, deep PNNs promise substantial benefits in speed and efficiency~\cite{markovic2020physics}, and may ultimately show scaling behavior similar to digital architectures.

Shifting from digital processors to analog physical systems brings new fundamental challenges. Of particular relevance is how to tame them to perform desired computations. Unlike DNNs, physical systems are often nondifferentiable, difficult to model accurately, and subject to uncertainty and noise - challenges that are escalated in deep, cascaded PNNs. As such, a strong wave of interest has emerged recently in training these PNNs~\cite{momeni2025training}. Conventional in silico training suffers from the simulation-reality gap~\cite{shen2017deep}. Hybrid approaches\cite{spall2022hybrid}, such as physics-aware training (PAT), mitigate this by interleaving physical forward passes with digital backpropagation (BP)~\cite{wright2022deep}. While generally powerful, they are limited in efficiency and adaptability, and remain dependent on digital models. Direct feedback alignment (DFA) is model-free because it exploits random projection rather than actual gradients for error backpropagation, but usually compromises performance~\cite{filipovich2022silicon,nakajima2022physical}. Another branch is physical local learning (PhyLL)~\cite{momeni2023backpropagation}, inspired by forward-forward algorithm~\cite{hinton2022forward,oguz2023forward}, which is widely compatible with physical systems but can be data-inefficient and memory-intensive due to necessitating multiple forward evaluations with contrastive samples. Many other methods \cite{hermans2015trainable,skalli2025model,pai2023experimentally,zhou2021large,gu2020flops,bandyopadhyay2024single,lopez2023self,xue2024fully,xu2024perfecting,cin2025training}, including equilibrium propagation~\cite{scellier2017equilibrium,stern2021supervised}, have also been explored, continuously expanding the diversity of unconventional PNN training methods.
Nevertheless, simultaneously achieving scalable, widely-compatible, and high-performance training of deep PNNs with minimal digital dependence remains challenging.

Here we introduce the physical information bottleneck (PIB), a training framework that addresses all above challenges by integrating information theory with local learning. PIB allocates an explicit training objective to each physical computing unit (Fig.~\ref{fig1}C), encouraging the distillation of task-relevant input features while filtering out irrelevant information. 
Consequently, all units jointly contribute to an optimized deep PNN with emergent capabilities. 
This allows us to optimize a unit of interest locally and independently, without the need of executing the entire network, storing all intermediate features and gradients, or waiting for sequential error backpropagation across layers. It also eliminates auxiliary digital models or lengthy contrastive measurements, thereby facilitating efficient training across diverse systems. We demonstrate the universality by training deep PNNs built on electronic memristor and free-space optical computing platforms, representing two broad categories of physical computing systems with known and unknown properties, respectively~\cite{momeni2025training}.

\section*{Results} 
\noindent\textbf{Local physical information bottleneck.} Any physical system can be viewed as a computing unit if its internal state dynamics maps external stimuli to task-favor outputs (Fig.~\ref{fig1}A). This viewpoint has motivated numerous physical computing paradigms across diverse fields~\cite{sebastian2020memory,wetzstein2020inference,shastri2021photonics,zhou2022photonic,torrejon2017neuromorphic,grollier2020neuromorphic,stern2020supervised,berloff2017realizing}, which can be broadly classified by their correspondence to digital operators (Fig.~\ref{fig1}A, right panel)~\cite{momeni2025training}. Isomorphic units are engineered to directly emulate specific mathematical operations in digital computers, such as matrix-vector multiplication (MVM). In contrast, broken-isomorphism units (non-isomorphic) leverage the native, often more complex or even unknown dynamics of physical systems for computation. When combined with a trainable readout, this broken-isomorphism approach echoes the framework of physical reservoir computing~\cite{tanaka2019recent}. These physical units, together with emerging theoretical frameworks~\cite{gallicchio2017deep,jaeger2023toward,lopez2023self}, now provides the foundation to scale up to deep PNNs (Fig.~\ref{fig1}B)~\cite{wright2022deep,momeni2023backpropagation,nakajima2022physical,bandyopadhyay2024single}.

\begin{figure*}[!htp]
  \centering{
  \includegraphics[width = 1.0\linewidth]{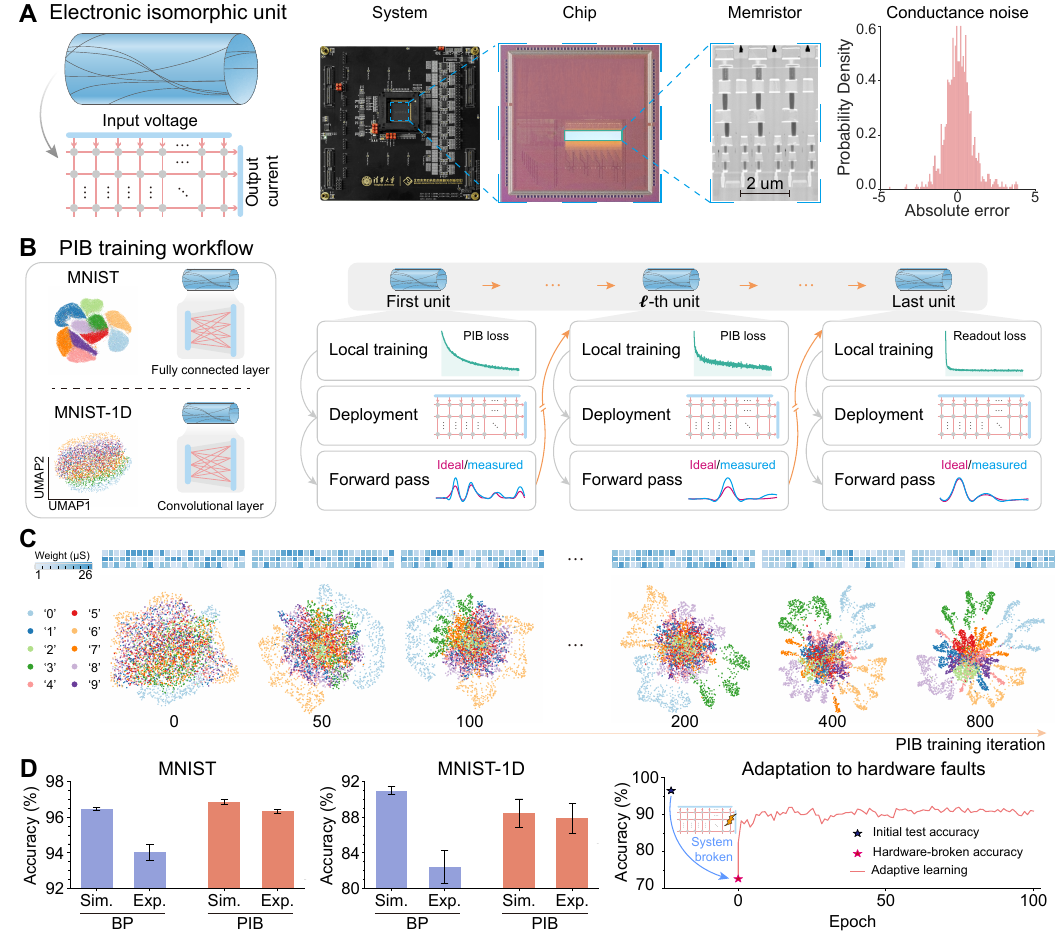}
  } 
    \caption{\noindent\textbf{Deep PNNs with isomorphic computations based on electronic memristors.} (\textbf{A}) The unit implements matrix-vector multiplication (MVM) via memristive crossbars. Panels show the packaged system, chip micrograph, and cross-sectional transmission electron microscopy image of one-transistor-one-resistor cells. The histogram shows shot-to-shot noise at 23.4~$\mu$S. (\textbf{B}) PIB training workflow. (Right) Different layers are used for datasets of varying complexity (visualized via UMAP). (Left) PIB training cycle includes local optimization, experimental deployment, and the forward pass to subsequent units. (\textbf{C}) Training dynamics showing weight evolution (top) and UMAP feature separation (bottom) over iterations. (\textbf{D}) System performance. (Left) Experimental PIB accuracy compared with in silico training benchmarks (mean $\pm$ s.t.d. across six runs). (Right) Demonstration of PIB's adaptation to severe hardware faults (broken analog-to-digital converter).
  }
  \label{fig2}
\end{figure*}

In a deep PNN, a stimulus carrying input data is progressively transformed into a desired output - such as a probability distribution - through hierarchical processing, where each unit produces intermediate feature representations that drive the next (Fig.~\ref{fig1}B). The PIB framework enables such hierarchical learning by optimizing each unit with a local objective inspired by the information bottleneck principle~\cite{tishby2000information,tishby2015deep}. Here, each unit is treated as an information channel: the output feature~$Z^{\ell}$ of the $\ell-\mathrm{th}$ unit is trained to retain relevant information for downstream tasks (e.g., predicting a label~$Y$) while suppressing redundant information from the original input~$X$ (Fig.~\ref{fig1}C). Such objective is formulated as maximizing a Lagrangian function~$\mathcal{L}(Z^{\ell},\beta) = I(Y;Z^{\ell}) - \beta I(X;Z^{\ell})$, where~$I(\cdot;\cdot)$ denotes mutual information and~$\beta$ is a Lagrange multiplier (Methods). We further instantiate PIB in Fig.~\ref{fig2}B. 

While information bottleneck theory has provided profound insights for interpreting digital DNNs~\cite{bang2021explaining}, we transform it from a passive interpretation method into an active training objective. We further harness a matrix-based formulation that overcomes the computational intractability of estimating mutual information in high-dimensional spaces~\cite{giraldo2014measures}. Such generalization establishes PIB as a practical training framework for experimental physical systems, enabling us to optimize deep physical computing layers, each governed by its own explicit matrix-based information bottleneck (detailed in Supplementary Note 1). 
Crucially, training a unit with PIB requires only local output measurements and the provided global input~$X$ and target~$Y$. This eliminates the end-to-end gradient flow required by previous deep PNN approaches and consequently reduces the digital memory footprint. The locality and information-theoretic foundation of PIB directly underpins the high training efficiency, noise robustness, and generalization capabilities demonstrated below.

\vspace{0.1cm}

\noindent\textbf{Memristor-based deep PNNs.}
First, we apply PIB to train deep PNNs built on an electronic isomorphic unit (Fig.~\ref{fig2}A). Our memristor-based platform executes parallel and reconfigurable in-memory MVM at a scale up to 1,024 $\times$ 128 (Methods). 
In experiments, inputs and weights are encoded as voltages and conductance states, respectively. After physical evolution, the output currents are naturally the computed results, mediated by Ohm's law and Kirchhoff's law~\cite{liu2025memristor}. While efficient, the intrinsic temporal fluctuation of conductance is one critical challenge in such analog systems (Fig.~\ref{fig2}A and Supplementary Note 3A). PIB training workflow accommodates these intrinsic fluctuations (Fig.~\ref{fig2}B). We construct deep PNNs including multilayer perceptrons and convolutional neural networks for tasks of varying complexity (Fig.~\ref{fig2}B, left). The process begins by training the first unit with its theoretical model using a PIB objective, calculated efficiently via the matrix-based mutual information between the unit's output feature~$Z^1$ and the global input~$X$ and target~$Y$.
Once these weights are deployed onto the chip, we capture the noisy experimental output features and utilize them as inputs for optimizing the subsequent unit. This cycle effectively incorporates hardware nonidealities into the learning process, without requiring specialized hardware improvements, noise-aware modeling, or global backpropagation. Thus, while individual units are subject to the simulation-reality gap, the deep PNN achieves global resilience through this cascaded, local experimental feedback. Moreover, the self-contained locality allows any physical unit to be trained in isolation without forwarding through the entire network or contrastive measurements, thereby granting our approach enhanced efficiency, experimental simplicity, and adaptability (Supplementary Note 2).

To visualize the training effects, we project the features preceding the final layer into a two-dimensional space via UMAP (Uniform Manifold Approximation and Projection)~\cite{mcinnes2018umap} (Fig.~\ref{fig2}C). A clear trend of clustering is evident with longer training, validating PIB as an effective objective. In Fig.~\ref{fig2}D, we benchmark classification performance against end-to-end BP (in silico training). For MNIST~\cite{lecun1998mnist}, PIB-based PNNs achieve average accuracies of 96.3\% and 96.8\% in experiments and simulations, respectively. In contrast, in silico training suffers a notable accuracy drop from 96.4\% to 94.0\% when deployed onto the hardware. For the challenging MNIST-1D~\cite{pmlr-v235-greydanus24a}, this degradation is more severe (8.5\%), while PIB maintains similar performance in both simulations and experiments, approaching the upper bound set by end-to-end BP in simulation (See Supplementary Note 3 for architecture and energy consumption details).

Beyond intrinsic noise, practical in-memory computing systems are vulnerable to component-level faults. Such corruptions can directly degrade deep PNN performance.
In our memristor device, an analog-to-digital converter for current readout is broken during the operation (Methods). This anomaly causes a catastrophic accuracy drop of $>$20\% on MNIST classification. Remarkably, the PIB workflow enables dynamic updates of the subsequent layers using these corrupted signals, eventually restoring performance from 73.5\% to over 90.0\% (Fig.~\ref{fig2}D). While retraining is required virtually for all approaches in such scenarios, PIB incurs minimal overhead by eliminating auxiliary networks and reducing memory footprint, therefore facilitating the robust, hardware-efficient usage of physical computing systems.

\vspace{0.1cm}

\begin{figure*}[!htp]
  \centering{
  \includegraphics[width = 1.0\linewidth]{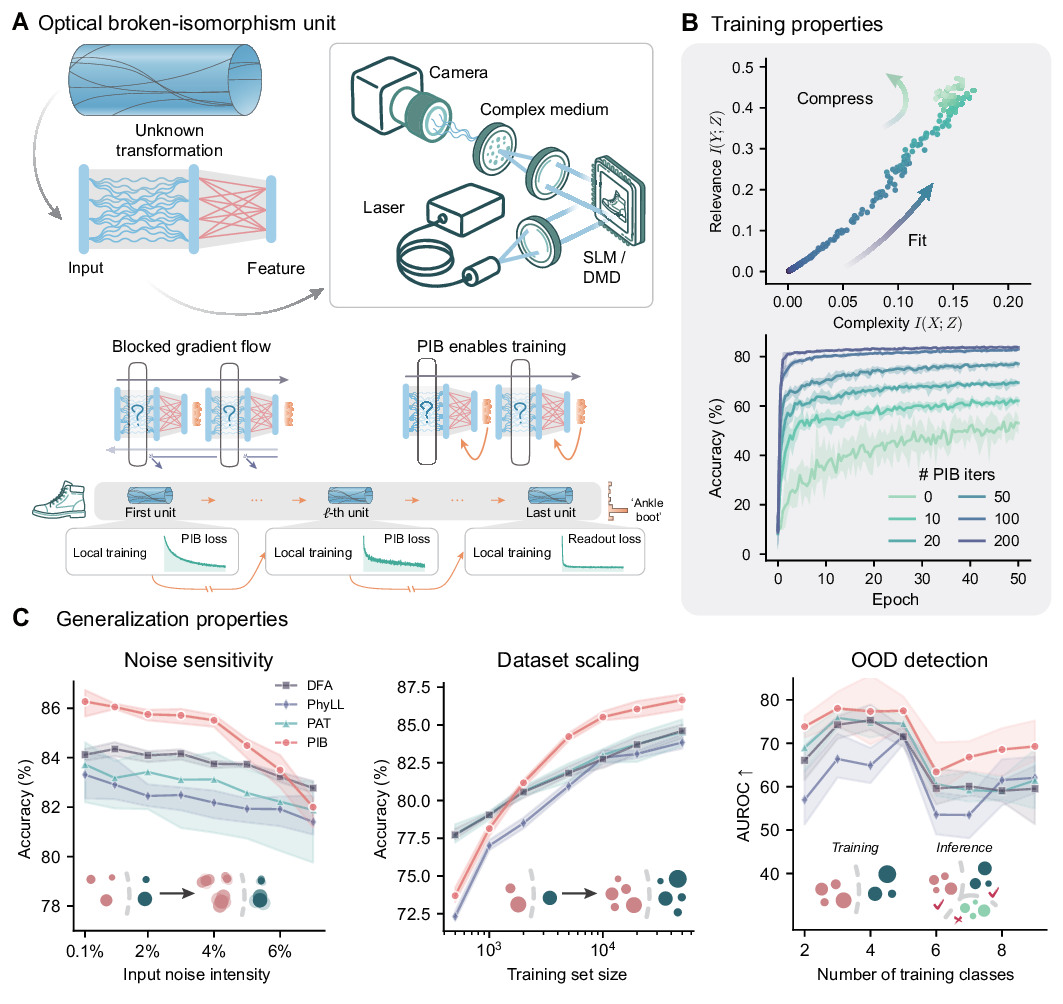}
  } 
    \caption{\noindent\textbf{Deep PNNs with broken-isomorphism computations based on optics.} (\textbf{A}) An optical processor maps encoded input fields to speckle features through multiple scattering. This unknown transformation breaks gradient flow, preventing standard global backpropagation. PIB optimizes each unit with a local objective, enabling unit-wise training without transmission-matrix retrieval or differentiable surrogates. (\textbf{B}) Training dynamics on Fashion-MNIST. Information plane trajectories exhibit a ``fit-then-compress" behavior ($I(Y;Z)\uparrow$ then $I(X;Z)\downarrow$). (Bottom) Readout training curves (x-axis: Epoch) show that increasing PIB iterations (color gradient) accelerates convergence and improves final accuracy. (\textbf{C}) PIB improves generalization over gradient-based (PAT), error-transport (DFA), and contrastive (PhyLL) baselines in noise robustness, data efficiency, and Out-of-Distribution (OOD) detection. Shaded regions denote multiple runs.
  }
  \label{fig3}
\end{figure*} 
\noindent\textbf{Optics-based deep PNNs.} 
We next extend PIB to deep PNNs based on an optical broken-isomorphism unit (second class in Fig.~\ref{fig1}A). The optical unit utilizes multiple light scattering in a disordered medium, where coherent light input is mixed into a speckle intensity pattern. Such phenomenon is ubiquitous across complex wave systems~\cite{rotter2017light}, enabling classical/quantum optical reservoir and kernel computing with a trainable layer ~\cite{gigan2022imaging}. Although the scattering process admits a well-defined model as a random projection, unlike memristive unit, identifying its exact physical parameters can be experimentally cumbersome. This black-box, non-differentiable transformation makes standard global backpropagation impractical. In experiments, we leverage the scattering process and photodetection as a nonlinear feature generator using two setups: a commercial Optical Processing Unit (OPU)~~\cite{brossollet2021lighton} and a custom-built optical setup, without retrieving the transmission matrix or its derivatives (Fig.~\ref{fig3}A). As in the memristor experiments, PIB optimizes each layer with the local objective $\mathcal{L}(Z^{\ell}, \beta)$, computed from the measured response statistics. We scale this unit into a deep PNN with multiple units, each comprising the fixed random projection followed by a trainable digital layer, and we train the model block-wise with PIB (Methods).

To visualize how PIB operates on this broken-isomorphism system, we monitor its matrix-based information plane dynamics (Fig.~\ref{fig3}B). We track $I\left(X;Z^{\ell}\right)$, as a measure of feature complexity, and $I\left(Y;Z^{\ell}\right)$, as a measure of target relevance. In the early phase, PIB primarily increases $I\left(Y;Z^{\ell}\right)$, indicating that the optics-based unit captures informative features. Subsequently, $I\left(X;Z^{\ell}\right)$ decreases while $I\left(Y;Z^{\ell}\right)$ remains high, consistent with compression of task-irrelevant variability while preserving predictive information. This ``fit-then-compress" trajectory is characteristic of the information bottleneck principle (more evident in Supplementary Note 3)\cite{tishby2015deep}. On Fashion-MNIST~\cite{xiao2017fashion}, this concentration of task-relevant information translates into improved performance. After PIB optimization, we freeze the optical units and probe the downstream performance with a linear classifier. The test accuracy increases with PIB iterations per unit, training converges faster, and the final accuracy distribution narrows for repeated runs. The optical deep PNN reaches up to 86.4\% with the OPU (binary encoding, full dataset) and 93.4\% test accuracy with the SLM-based setup (8-bit encoding, subset; Methods). 

The way PIB distills task-relevant information further suggests enhanced generalization. We therefore conduct three generalization tests (Fig.~\ref{fig3}C). Compared to alternative approaches, PIB shows higher accuracy under noise level up to $5\%$, better sample efficiency in data-scarce regimes, and higher out-of-distribution (OOD) detection rates for unseen classes. Collectively, these experimental results support that PIB provides a practical and effective route to training deep PNNs based on non-differentiable broken-isomorphism units, delivering good performance and enhanced robustness to noise, data scarcity and class shift in this optical setting.

\begin{figure*}[!htp]
  \centering{
  \includegraphics[width = 1.0\linewidth]{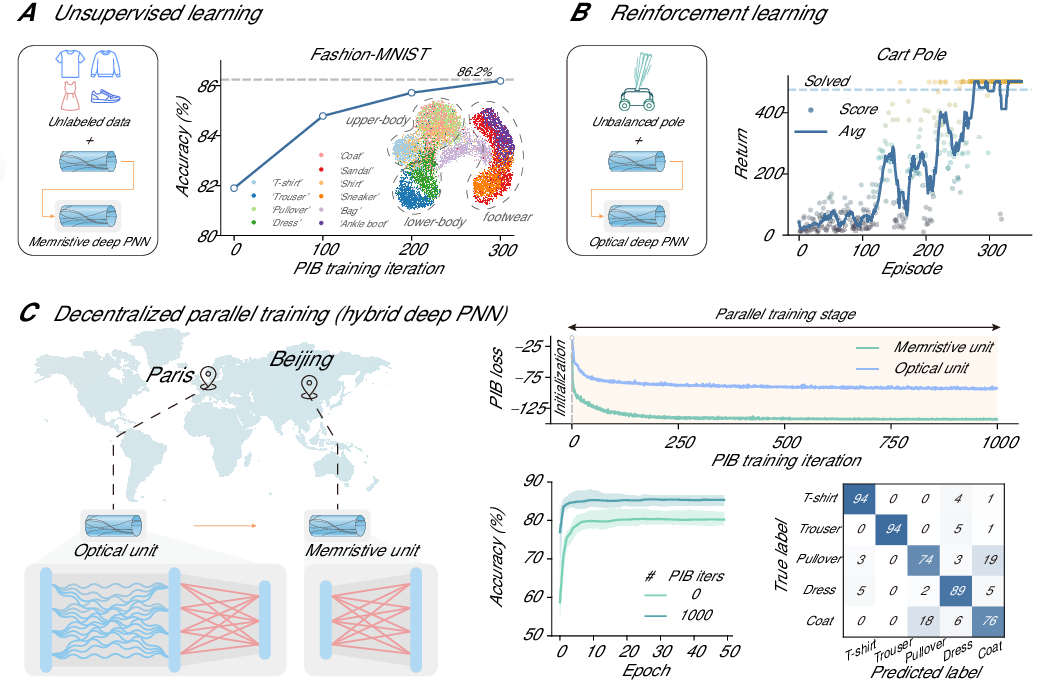}
  } 
    \caption{\noindent\textbf{Versatility of the PIB framework in unsupervised learning, reinforcement learning, and decentralized parallel training regimes.} (\textbf{A}) A memristor-based deep PNN is trained by PIB's unsupervised learning variant. Increasing PIB training iterations enhance feature's linear separability as shown by the improved accuracy of a downstream linear probe. (Inset) Semantically disentangled UMAP visualization. (\textbf{B}) Reinforcement learning on an optics-based deep PNN agent. The agent learns the CartPole-v1 control task via a local loss, a distinct departure from standard RL. By minimizing a composite local objective of PIB and temporal-difference Q-learning loss, episode return (Score) and its running average (Avg) increase effectively to reach the solved regime. (\textbf{C}) Decentralized parallel training of a hybrid optical-electronic deep PNN. Physical computing units are geographically distributed across two sites, and PIB allows for simultaneous, decoupled optimization using local computing resource. (Right) Parallel training loss curves for both units demonstrate convergence of respective PIB objectives without inter-site communication or synchronization. (Bottom) Downstream task results after parallel training.
  }
  \label{fig4}
\end{figure*} 

\vspace{0.1cm}

\noindent\textbf{Scaling deep PNNs to versatile scenarios.} Finally, to illustrate the potential of PIB for broader scenarios, we demonstrate several variants in Fig.~\ref{fig4}. In unsupervised learning or self-supervised learning, the deep PNN should extract features from unlabeled data. We modify PIB objective to $\mathcal{L}(Z_1,Z_2,\beta) = I(Z_1;Z_2) - \beta[ I(X_1;Z_1)+I(X_2;Z_2)]$, where~$X_{1,2}$ represent two random augmented views of the same input~$X$, and~$Z_{1,2}$ are the corresponding features~\cite{liu2024learning}. This objective trains the unit to extract view-invariant features by maximizing~$I(Z_1;Z_2)$, while discarding view-specific information. Thus, negative data-pairs are avoided as in contrastive self-supervised learning~\cite{chen2020simple} (Methods). Remarkably, the UMAP visualization (Fig.~\ref{fig4}A, inset) reveals that the memristive deep PNN learns to semantically distinguish upper- and lower-body clothing from footwear without using any labels. This representational quality is quantitatively confirmed by the standard linear evaluation protocol. As shown in Fig.~\ref{fig4}A, the downstream Fashion-MNIST accuracy increases monotonically with PIB training iterations.

We next extend PIB to reinforcement learning (RL) on the CartPole-v1 using an optics-based deep PNN. Here, the learning signal is inherently noisy and non-stationary~\cite{mnih2015human}, and such fluctuations are compounded by measurement noise and drift in the physical hardware. A second practical challenge is the latency mismatch between fast optical propagation and comparatively slow device reconfiguration, which makes tightly-synchronized, end-to-end RL updates inefficient and can leave the deep PNN idle while awaiting for global feedback. To address both issues, we develop a local training protocol that combines a temporal-difference (TD) Q-learning loss with a statistics-based PIB for each unit, $\mathcal{L}^{\ell}_{\mathrm{local}}=\mathcal{L}^{\ell}_{\mathrm{PIB}}+\lambda\,\mathcal{L}^{\ell}_{\mathrm{TD}}$ (Methods). Here, the PIB term is computed directly from the measured responses of the current batch, providing an immediate guiding signal that extracts relevant information while filtering out both hardware noise and RL stochasticity. This formulation enables asynchronous updates by overlapping the idle periods of the physical hardware and with the Q-learning computation, effectively masking the latency mismatch. Consequently, the PNN agent reliably reaches the solved regime (Fig.~\ref{fig4}B), demonstrating that PIB remains effective under hardware noise and high-variance RL targets.

By virtue of its local, self-contained nature, PIB is particularly suitable for decentralized parallel training. Unlike end-to-end methods requiring sequential layer updates, PIB allows simultaneous training of multiple physical units, each optimized by its own objective. In experiments, we construct a heterogeneous deep PNN with an optical scattering unit in Paris and an electronic memristive unit in Beijing (Fig.~\ref{fig4}C). Following a single initialization pass to distribute inputs to all hidden units, the network is trained fully in parallel (rather than in a cascade as in Figs.~\ref{fig2},~\ref{fig3}), without requiring synchronized updates. Each site therefore trains its unit using only local computing resources (Methods). This distributed framework avoids global memory locking, reduces the computational load per node, shortens wall-clock training time via parallelization, and provides inherent protection of model privacy and physical hardware intellectual property~\cite{warnat2021swarm}.

At present, the main limitation of PIB is the reliance on digital auto-differentiation to retrieve gradients of the training parameters (Supplementary Table 1). However, this constraint can be alleviated by integrating PIB with other surrogate gradients methods, such as DFA. We experimentally demonstrate this integration in Supplementary Note 4, showing that PIB can still serve as an effective information-theoretic objective to guide the optimization of internal physical parameters via DFA.

\section*{Discussion}
\noindent{PIB} rests on three core principles: (i) it is a general framework compatible with multiple physical platforms, encompassing both isomorphic and broken-isomorphism physical computing units; (ii) it is grounded in matrix-based information theory, offering a conceptually intuitive and computationally tractable approach to evaluate entropy in high-dimensional spaces; and (iii) it is ideally streamlined for experimental implementation, requiring neither auxiliary digital models nor contrastive measurements. Across tasks of varying complexity, PIB approaches the performance upper bound of ideal BP in simulation, and frequently outperforms existing deep PNN training methods (Fig.~\ref{fig3})~\cite{wright2022deep,nakajima2022physical,filipovich2022silicon,momeni2023backpropagation}. Beyond competitive accuracy, its locality and information-theoretic foundation yield several fundamental hardware advantages as detailed in Supplementary Note 2.

Many other physical computing systems can be employed to construct noise-robust deep PNNs using this approach. For isomorphic computing units like photonic matrix-vector multipliers~\cite{shen2017deep,feldmann2021parallel,tait2017neuromorphic}, PIB reduces the simulation-reality gap by progressively training each layer with experimental feedback. For broken-isomorphism systems, such as acoustic wave reservoirs~\cite{meffan2023non}, magnetic skyrmions~\cite{prychynenko2018magnetic}, or active matter~\cite{wang2024harnessing}, PIB empowers them to explore hierarchical processing capabilities augmented by trainable intermediate layers beyond reservoir computing~\cite{appeltant2011information,tanaka2019recent,gallicchio2017deep}. When scaling to highly complex tasks, PIB's performance may lag behind end-to-end BP from the known trade-offs between locality and global optimization. However, recent digital benchmarks suggest that this gap is narrowing~\cite{belilovsky2019greedy}. We thus believe that this tradeoff is justified, considering the significant physical hardware benefits (Supplementary Table 1).
To achieve scalable performance with deep PNNs, future work could incorporate key inductive biases inherent to the physical computing unit (see expanded discussions in Supplementary Note 5).

Broadly, as the information bottleneck principle sees emerging applications across neuroscience~\cite{sachdeva2021optimal}, complex systems~\cite{murphy2024information}, and foundation models~\cite{shani2025tokens}, PIB represents a unique synergy of neuromorphic physical computing and statistical machine learning. It provides a route to effectively train future large-scale, energy-efficient, and ultrafast deep PNNs, ultimately enabling the investigation of their scaling laws.

\vspace{0.1cm}

\vspace{0.5cm}

\begin{footnotesize}

\noindent \textbf{Data and code availability}: 
The data and codes that support
the results within the paper are
available at 
\href{https://doi.org/10.5281/zenodo.18450060}{Zenodo}.
\vspace{0.1cm}

\noindent \textbf{Acknowledgements}: The authors thank Yanze Zhou for his help in memristive experiments. H.Wang, X.L., H.Z., and J.T. acknowledge support from the National Natural Science Foundation of China (Grants 623B2064, 62404122, 625B2105, and 92264201, respectively). Z.W. acknowledges support from PSL Research University. J.H. acknowledges the startup fund from the University of Hong Kong. X.F. acknowledges support from the Beijing Natural Science Foundation (Grant JQ23021). S.G. is a member of the Institut Universitaire de France. This work was supported by the Shenzhen Science and Technology Program (No. CJGJZD20240729141102004), and the Tsinghua University (Department of Precision Instrument)-North Laser Research Institute Co., Ltd Joint Research Center for Advanced Laser Technology (20244910194). \vspace{0.1cm}

\noindent \textbf{Author contributions}: H.Wang and Z.W. conceived the study and performed the theoretical and numerical simulations. H.Wang and X.L. conducted the memristive experiments under the supervision of J.T. and H.Wu.; H.Z. contributed to the characterization of the memristive platform; Z.W. carried out the optical experiments. J.J. constructed the optical setup and assisted H.Wang with the optical experiments. J.T., H.Wu, S.G., and Q.L. supervised the project. H.Wang, Z.W., and J.H. wrote the manuscript with inputs from all authors.
\vspace{0.1cm}

\noindent \textbf{Competing interests}: The authors declare that they have no competing interests.
\end{footnotesize}

\newpage
\renewcommand{\bibpreamble}{
$^\dagger$These authors contributed equally to this work.\\
$^\ast${Corresponding authors: \textcolor{magenta}{sylvain.gigan@lkb.ens.fr}, \textcolor{magenta}{qiangliu@tsinghua.edu.cn}}\\
}

\bibliographystyle{naturemag}
\bibliography{ref}

\end{document}